\documentclass{article} % For LaTeX2e

\usepackage{iclr2027_conference,times}

\ifdefined\SWEGameMathCommandsLoaded
  \expandafter 
\fi
\def\SWEGameMathCommandsLoaded{1}

\usepackage{amsmath,amsfonts,bm}

\def\eqref#1{equation~\ref{#1}}
\def\1{\bm{1}}

\DeclareMathAlphabet{\mathsfit}{\encodingdefault}{\sfdefault}{m}{sl}
\SetMathAlphabet{\mathsfit}{bold}{\encodingdefault}{\sfdefault}{bx}{n}

\ifcsname argmax\endcsname\else

\fi
\ifcsname argmin\endcsname\else

\fi

\ifcsname sign\endcsname\else

\fi
\ifcsname Tr\endcsname\else

\fi

\usepackage{amssymb}
\usepackage{graphicx}
\usepackage{booktabs}
\usepackage{amsmath}
\usepackage{tabularx}
\usepackage{array}
\usepackage{multirow}
\usepackage{wrapfig}
\usepackage{caption}
\usepackage{capt-of}
\usepackage{url}
\usepackage{hyperref}
\usepackage{float}

\usepackage[most]{tcolorbox}
\usepackage{listings}
\tcbuselibrary{listings,breakable}

\newtcblisting{taskbox}[1][]{
    enhanced,
    listing only,
    colback=gray!3,
    colframe=black!70,
    colbacktitle=black!75,
    coltitle=white,
    fonttitle=\small\bfseries,
    boxrule=0.8pt,
    arc=1.5mm,
    outer arc=1.5mm,
    left=2mm,
    right=2mm,
    top=1.5mm,
    bottom=1.5mm,
    boxsep=1mm,
    before skip=6pt,
    after skip=8pt,
    listing options={
        basicstyle=\ttfamily\footnotesize,
        columns=fullflexible,
        keepspaces=true,
        showstringspaces=false,
        breaklines=true
    },
    #1
}

\providecommand{\bench}{SWE-Game}
\title{SWE-Game: Can Coding Agents Build the Games We Want?}

\iclrfinalcopy
\author{%
\parbox[t]{\dimexpr\textwidth-2\tabcolsep\relax}{%
\centering
\bfseries
Xiaoyu Chen\textsuperscript{1}\thanks{These authors contributed equally to this work.}\quad
Lai Wei\textsuperscript{1,2}\footnotemark[1]\quad
Jin Wang\textsuperscript{1}\footnotemark[1]\quad
Xiangyu Zou\textsuperscript{4}\quad
Ruochen Fan\textsuperscript{1}\quad
Enze Luo\textsuperscript{1}\\[4pt]
Mingzhe Yao\textsuperscript{1}\quad
Jiahui Zhu\textsuperscript{6}\quad
Yuhua Wen\textsuperscript{5}\quad
Linghe Kong\textsuperscript{1}\quad
Weiran Huang\textsuperscript{1,3}\thanks{Corresponding author.}\\[6pt]
\normalfont\small
\textsuperscript{1}\,Shanghai Jiao Tong University\quad
\textsuperscript{2}\,Zhongguancun Academy\\[2pt]
\textsuperscript{3}\,Shanghai Innovation Institute\quad
\textsuperscript{4}\,Shenzhen University\\[2pt]
\textsuperscript{5}\,Beijing University of Posts and Telecommunications\quad
\textsuperscript{6}\,Elbetech Technology\\[4pt]
\texttt{weiran.huang@outlook.com}
}%
}

\begin{document}

\maketitle
\lhead{Preprint}
\begingroup
\makeatletter
\def\@captype{figure}
\makeatother

\noindent
\begin{minipage}{\linewidth}
    \centering

    \includegraphics[
        width=\linewidth,
        trim=30.5bp 264.5bp 58.3bp 97.6bp,
        clip
    ]{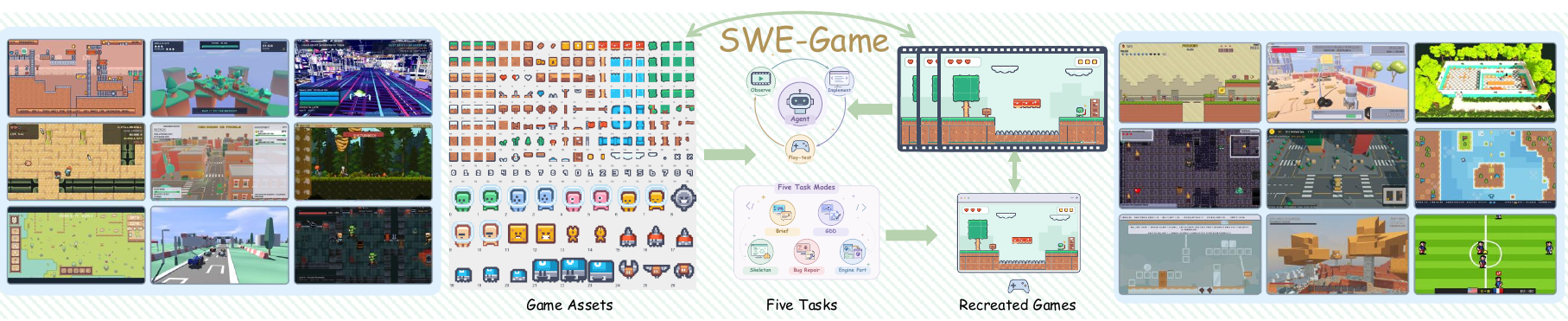}

    \caption{%
        \textbf{\bench{}: from reference gameplay to playable games.}
        The center illustrates an agent using supplied assets and
        gameplay demonstrations to recreate the target appearance
        and behavior through coding and play-testing.
        Five tasks vary the requirements and starting project,
        covering construction, repair, and cross-engine porting.
    }
    \label{fig:teaser}
\end{minipage}

\par
\endgroup
\medskip

\begin{abstract}
We introduce \bench{}, a benchmark of 247 tasks grounded in 41 executable reference Godot games spanning 13 gameplay categories in 2D and 3D. Five task types cover development from a brief, implementation from a game design document, skeleton completion, repair of 83 injected-fault cases, and Godot-to-Unity porting. Reference materials specify the intended gameplay, while a shared instrumentation interface lets evaluator-owned drivers and probes execute actions and observe independently implemented games. Evaluation combines engine-state checks, certified reference-input replay, and agent-authored feature demonstrations to assess mechanic correctness, demonstrated playability, and behavioral restoration and preservation after repairs. Game-specific vision-language rubrics separately assess presentation. Across six models, Opus5 achieves the highest overall score in all five task types. Best overall scores remain below 60 out of 100 across the three construction tasks, with Brief-to-Game reaching 50.38. Analysis of reviewed submissions identifies requirement omissions and gameplay logic errors as predominant implementation problems. On human-labeled behaviors from 100 agent-built games, executable checks achieve 92.59\% balanced accuracy, compared with 78.41\% for a video-based VLM judge. Rubric-based visual scores reach a Spearman correlation of 0.829 with human ratings of 200 gameplay clips. Together, these results characterize current agent capabilities across game-development activities and support combining runtime evidence with visual assessment.
\end{abstract}

\section{Introduction}
\label{sec:introduction}

Game development requires coding agents to coordinate code, assets, and engine behavior into an interactive experience. Advances in game generation support increasingly complete development workflows \citep{luo2026gamecraft,huang2026gui}, while coding agents provide repository navigation, editing, and execution capabilities for iterative implementation and testing \citep{yang2024swe,wang2025openhands,chi2026gamedevbench}. In practice, developers may begin with prepared characters, environments, and props, together with a gameplay demonstration of the desired result \citep{earle2025dreamgarden,yin2026autoue}. Agents must translate these materials into mechanics, progression, and presentation, then preserve the intended gameplay as projects are completed, repaired, or migrated.(Figure~\ref{fig:teaser})

Evaluating these activities requires connecting the intended gameplay to evidence of its implementation. A gameplay demonstration communicates movement, timing, and feedback, but the implementation must reproduce the interactions that generate these observations \citep{liu2026mage}. A platformer may display spikes correctly while allowing the player to pass through unharmed. Reaching a victory screen also provides incomplete evidence when required interactions have been bypassed \citep{peng2026proxywar}. Evaluation therefore needs explicit behavioral targets and execution evidence for the required interactions, complemented by visual assessment of how the game communicates those behaviors to the player.

Existing game benchmarks differ both in the development activities they cover and in how they assess the resulting games. GameDevBench and JamBench cover project editing, generation, and completion \citep{chi2026gamedevbench,sun2026jamer}, while GameEngineBench examines C++ integration within Unreal Engine projects \citep{la2026gameenginebench}. For complete-game construction, GameCraft-Bench uses replayed demonstrations and multimodal rubrics \citep{luo2026gamecraft}. WebGameBench evaluates browser games through runtime interaction \citep{zhang2026webgamebench}, and concurrent GameLogicBench tests mechanics through deterministic state and event assertions \citep{che2026gamelogicbench}. We focus on connecting task specification and evaluation across construction, maintenance, and migration, so that the requested gameplay determines the behaviors checked in the output.

\begin{figure}[t]
    \centering
    \includegraphics[width=\linewidth,trim=60bp 3bp 57bp 21bp,clip]{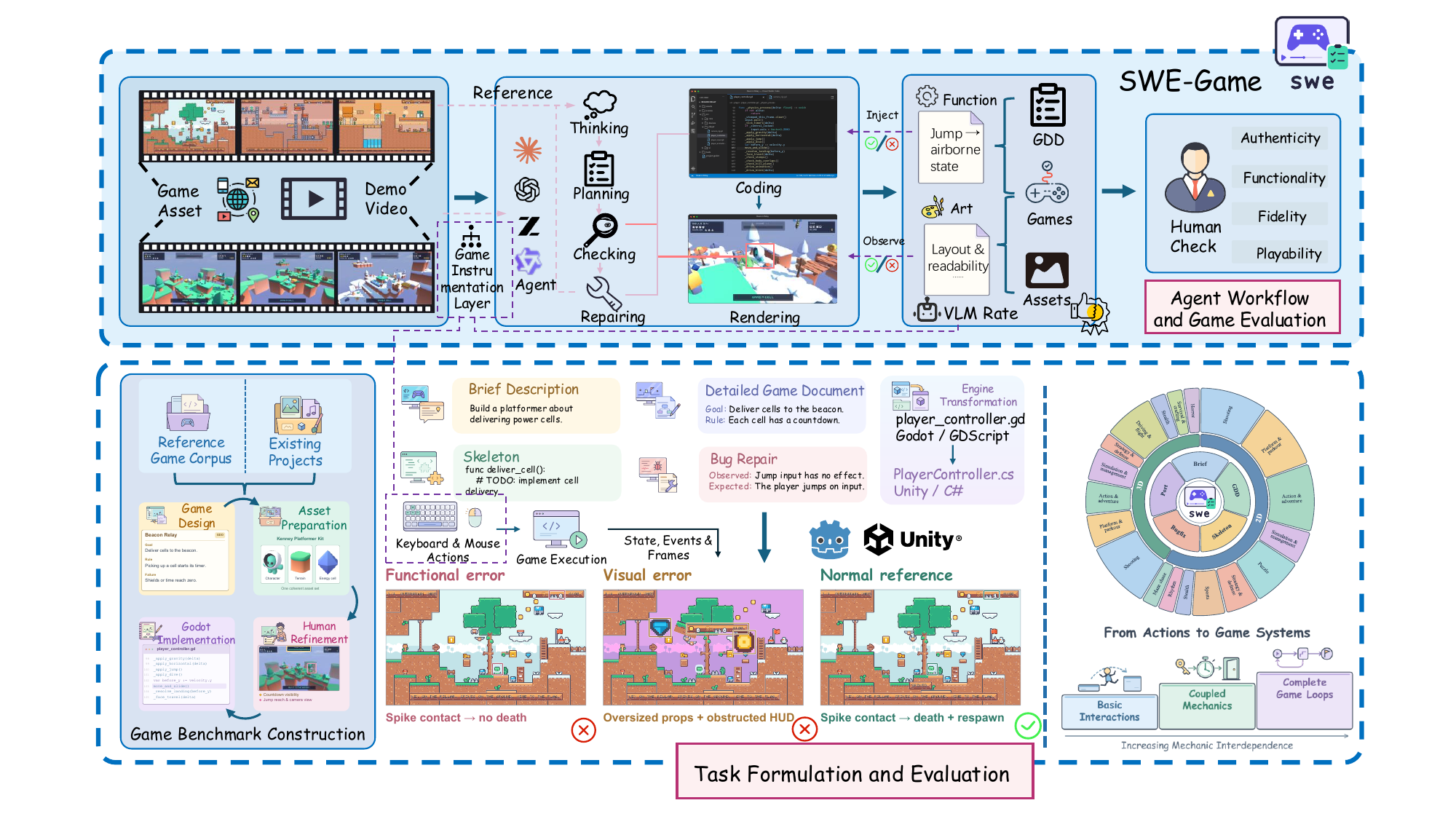}
    \caption{\textbf{Overview of \bench{}.} \textbf{Top:} agent development from reference materials through coding and execution, followed by assessment using functional tests, visual rubrics, and human validation. Visual rubric scoring applies to construction tasks. \textbf{Bottom:} reference-game construction, five development tasks and their evaluation, and game coverage. The gameplay panels use constructed examples to illustrate functional failure, visual failure, and correct reference behavior.}
    \label{fig:overview}
\end{figure}

We introduce \bench{}, a benchmark that establishes this connection through executable reference games. Task-specific combinations of videos, assets, requirements, and project code communicate the intended experience, while the executable references establish concrete expectations for evaluation. From 41 reference Godot games spanning 13 gameplay categories in 2D and 3D, we derive 247 tasks covering brief-to-game generation, GDD-to-game implementation, skeleton completion, bug repair, and Godot-to-Unity porting. Each non-repair task type contains 41 tasks, and the repair suite contains 83 cases. Across these tasks, requirements specify the behavior and content to reproduce or preserve while permitting alternative implementations (Figure~\ref{fig:overview}).

Evaluating these alternative implementations requires identifying equivalent gameplay roles across different object names, scene hierarchies, and code. Our \emph{game instrumentation interface} provides common access to playable scenes, input actions, semantic object roles, and relevant state fields. Submissions bind these roles to their own structures, enabling evaluator-owned drivers and probes to execute actions and inspect engine state. In the spike example, these bindings allow the evaluator to locate the player and hazard, observe contact, and check for the required health change or respawn.

Using these bindings, the evaluator runs local interaction tests and longer input replays to assess mechanics and progression against reference measurements and task requirements. Matched no-input controls check whether demonstrated completion depends on player actions. For repair, fixed scenarios run on the reference, faulty, and repaired versions to measure restoration and preservation, including faults spanning scripts, scenes, and resources. Porting uses a corresponding Unity execution contract. To assess how gameplay is communicated to players, game-specific vision-language rubrics evaluate presentation, visual feedback, and asset integration for construction tasks, complementing the functional evidence from execution.

We evaluate six models across the five task types. Opus5 achieves the highest overall score in each task, while the relative rankings of other models vary across development activities. Analysis of reviewed submissions identifies requirement omissions and gameplay logic errors as predominant implementation problems. We also quantify evaluator validity against human judgments: executable checks achieve 92.59\% balanced accuracy on labeled behaviors from 100 agent-built projects, compared with 78.41\% for a video-based VLM judge. Rubric-based visual scores reach a Spearman correlation of 0.829 with human ratings of 200 gameplay clips.

Our main contributions are:
\begin{itemize}
    \item \textbf{A benchmark grounded in executable reference games.} We construct 247 tasks across five development activities, using reference materials to specify the intended experience and executable games to establish behavioral targets.

    \item \textbf{Runtime evaluation across independent game implementations.} A shared instrumentation interface connects task requirements to input replay and engine-state observations, enabling functional checks across different project structures alongside visual assessment.

    \item \textbf{A systematic study of agent capabilities and evaluation validity.} We evaluate six models, analyze implementation problems across reviewed submissions, and measure the accuracy of functional judgments and the correlation of visual scores with human ratings.
\end{itemize}

\section{Related Work}

\textbf{Coding as an executable task.} Software engineering evaluation has shifted toward executable tasks centered on execution environments, interaction, and sustained engineering progress. SWE-bench catalyzed this shift by combining repository modifications with bug-fixing and regression tests\citep{jimenez2024swe}. Subsequent benchmarks, including CRUXEval\citep{gu2024cruxeval}, LiveCodeBench\citep{jain2025livecodebench}, and BigCodeBench\citep{zhuo2025bigcodebench}, further broadened evaluation coverage. This paradigm was further advanced by ProjectEval\citep{liu2025projecteval} and Terminal-Bench\citep{merrill2026terminal}: They introducing interactive assessment of complete projects and a unified environment–task–test protocol, respectively. More recently, SWE-Marathon has extended the paradigm to to long-horizon, project-level workflows \citep{desai2026swe}. Building on this line of work, SWE-Game defines task contracts, behavior-preservation checks, and explicit model–evaluator interaction protocols for different development paradigms.

\textbf{Game development benchmarks.} Recently, GameDevBench introduced game development into autonomous-agent evaluation\citep{chi2026gamedevbench}, extending software engineering benchmarks to interactive game construction. Since then, end-to-end benchmarks such as OpenGame\citep{jiang2026opengame}, WebGameBench\citep{zhang2026webgamebench}, and V-GameGym\citep{zhang2026v} have emerged, using multimodal models or agents to evaluate generated games. JAMER\citep{sun2026jamer} and GameXpert-Bench\citep{chen2026gamexpert} further broaden task coverage and introduce predominantly rule-based verification.Existing approaches, however, face two challenges. First, open-ended or narrowly defined tasks limit the assessment of agents’ multimodal understanding and system-building capabilities. Second, rule-based tests are often too coarse-grained to capture dynamic responses and overall logical consistency during interaction. SWE-Game is designed to address these limitations. It covers five development contracts over a shared project foundation, enables fine-grained runtime evaluation, and exposes the evidence underlying behavioral and perceptual outcomes rather than relying solely on static code inspection, coarse-grained comparisons, or model-judge scores.

\textbf{Interactive repair and diagnostic testing.}
PlayCoder employs active GUI testing, behavioral reporting, and iterative repair \citep{peng2026playcoder}. Play2Code integrates coding with GUI playtesting and introduces persistent memory \citep{huang2026gui}. AutoSD organizes debugging around hypotheses, experiments, and observations \citep{kang2025explainable}, while Causal Testing constructs minimally different executions to explain faulty behavior \citep{johnson2020causal}. These approaches guide revisions through execution feedback, but additional iterations need not improve outcomes. In mutation-test generation, scientific-debugging prompts required more iterations than ordinary iterative prompting while achieving similar success rates \citep{straubinger2025mutation}. This distinction motivates evaluating the resulting behavior against explicit requirements. GameGen-Verifier checks local gameplay keypoints through runtime state injection \citep{jia2026gamegen}. SWE-Game checks required interactions using evaluator-owned probes and input replays, with matched no-input controls testing action dependence. For repair, identical scenarios run on reference, faulty, and repaired projects to assess restoration and preservation.

\section{\bench{} Benchmark}
\label{sec:benchmark}

\bench{} evaluates whether coding agents can build, complete, repair, and port
games against fixed gameplay and visual targets. Its 247 tasks derive from 41
executable reference games (Figure~\ref{fig:overview});
Table~\ref{tab:benchmark_comparison} contrasts its inputs, task types, and
evaluation with existing game-development benchmarks.

\subsection{Reference Game Construction}
\label{sec:corpus}

\begin{figure}[t]
    \centering
    \includegraphics[width=\linewidth]{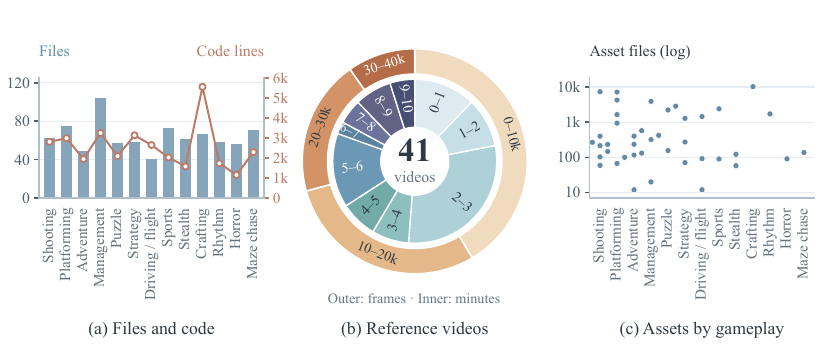}
    \caption{\textbf{Reference game statistics.} (a) Median numbers of Godot script and scene files (bars) and GDScript code lines (line) within each gameplay category. (b) Distributions of reference-video duration (inner ring, minutes) and frame count (outer ring). (c) Supplied asset-file counts by gameplay category, with one point per game on a logarithmic scale.}
    \label{fig:reference_statistics}
\end{figure}

The 41 reference games are Godot projects, 24 in 2D and 17 in 3D, spanning
platforming, shooting, puzzle, strategy, simulation, and other gameplay types.
Each begins from licensed public assets, such as Kenney packs, whose
characters, environments, and interface elements shape its theme. A game
design document (GDD) then fixes the objective, controls, core mechanics,
success and failure conditions, content scope, and the role of each asset.
Levels follow an introduce, vary, and combine progression: a mechanic is first
introduced, then used under changed conditions, and finally combined with
others. In Harvest Ledger, for example, planting, harvesting, and selling
establish a farming economy, after which resource gathering and livestock
compete for the player's time and budget. The GDD specifies such interactions
as action-dependent state changes with visual or audio feedback. Each design
is implemented in Godot scenes and scripts and revised through runtime checks
and human playtesting, with the GDD updated alongside the code. The finished
projects, GDDs, assets, and recorded gameplay form the reference materials for
task construction; Figure~\ref{fig:reference_statistics} summarizes their
sizes, video durations, and asset counts.

\begin{table}[t]
\centering
\caption{Comparison of Game-Development Benchmark Protocols}
\label{tab:benchmark_comparison}

\begingroup
\footnotesize
\setlength{\tabcolsep}{3pt}
\renewcommand{\arraystretch}{1.08}

\resizebox{\linewidth}{!}{%
\begin{tabular}{@{}llclllc@{}}
\toprule
\textbf{Benchmark}
& \textbf{Runtime}
& \textbf{Input}
& \textbf{Tasks}
& \textbf{Evaluation}
& \textbf{Scoring unit}
& \shortstack{\textbf{Det. runtime}\\\textbf{checks}} \\
\midrule

OpenGame~\citeyearpar{jiang2026opengame}
& Web
& T
& G
& VLM judge
& Game, dimension
& $\times$ \\

WebGameBench~\citeyearpar{zhang2026webgamebench}
& Web
& T
& G
& Agent judge
& Requirement
& $\times$ \\

V-GameGym~\citeyearpar{zhang2026v}
& Pygame
& T
& G
& VLM judge
& Game, dimension
& $\times$ \\

VERIGAME~\citeyearpar{jia2026gamegen}
& Web
& T
& G
& State inj., VLM
& Keypoint
& $\times$ \\

GameXpert-Bench~\citeyearpar{chen2026gamexpert}
& Web
& T, S
& G, O, R
& Runtime tests, human
& Event, assertion
& $\circ$ \\

GameCraft-Bench~\citeyearpar{luo2026gamecraft}
& Godot
& T, A
& G
& VLM judge
& Rubric item
& $\times$ \\

GameDevBench~\citeyearpar{chi2026gamedevbench}
& Godot
& T, S, A, I, V
& Patch
& Unit tests
& Operation, test
& $\checkmark$ \\

JamBench~\citeyearpar{sun2026jamer}
& Godot
& T, S, A
& G, C
& Engine verify
& Project, code, runtime
& $\checkmark$ \\

GameLogicBench~\citeyearpar{che2026gamelogicbench}
& Godot
& T, S, A
& Patch
& Engine assertions
& Scenario
& $\checkmark$ \\

\midrule
\textbf{SWE-Game (Ours)}
& \textbf{Godot, Unity}
& \textbf{T, S, A, V\textsuperscript{*}}
& \textbf{G, C, R, P}
& \textbf{Probes, replay, VLM}
& \textbf{Criterion, route, keypoint}
& $\checkmark$ \\
\bottomrule
\end{tabular}%
}

\par\vspace{2pt}
\parbox{\linewidth}{\footnotesize\raggedright
\textit{Input:}
T, text; S, source code; A, assets; I, images; V, video.
Input forms are listed across task modes; individual tasks may use a subset.
\textit{Tasks:}
G, generation; C, completion; O, optimization;
R, repair; P, porting.
Det. runtime checks refers to the objective runtime evaluation path:
$\checkmark$, deterministic; $\times$, non-deterministic;
$\circ$, track-dependent.
\textsuperscript{*}Video input is mode-dependent;
Bug Repair uses a faulty project and textual requirements and report, without a reference video.
\par}

\endgroup
\end{table}

\subsection{Task Formulation}
\label{sec:tasks}

Five task modes vary the specification and starting implementation supplied
for a reference game (Table~\ref{tab:swe_game_tasks}). Each mode has one task
per game except Bug Repair, which has 83 cases, for 247 tasks in total.

\paragraph{Construction.}
Three modes differ in how much of the design and implementation is given.
\emph{Brief-to-Game} supplies a short request and requires the agent to author
a GDD before building the game; \emph{GDD-to-Game} supplies the full design;
\emph{Skeleton Completion} supplies an unfinished project whose missing
functionality must be integrated with existing scenes and logic. All three
provide assets and a reference video, and the requirements state which design
choices remain open. Submissions also include feature demonstrations,
player-action sequences showing that required features work in the submitted
game.

\paragraph{Repair and porting.}
\emph{Bug Repair} supplies a project that launches but misbehaves, its
requirements, and a player-facing report that describes symptoms without
naming files or lines. Faults include state that is not reset across levels
and partially migrated mechanisms whose readers and writers disagree. They are
mixed with suspicious but harmless edits, and some require coordinated fixes
in several locations. \emph{Godot-to-Unity Porting} supplies the Godot source,
GDD, assets, reference video, and a target Unity interface, and requires a
buildable Unity project that preserves the controls, mechanics, and
progression of the original.

\begin{table}[t]
    \centering
    \caption{Task modes in \bench{}: agent inputs,
    required artifacts, and evaluation components.}
    \label{tab:swe_game_tasks}
    \renewcommand{\arraystretch}{1.15}
    \resizebox{\linewidth}{!}{
        \begin{tabular}{@{}llll@{}}
            \toprule
            \textbf{Task}
            & \textbf{Agent inputs}
            & \textbf{Required artifacts}
            & \textbf{Evaluation components} \\
            \midrule
            
            Brief-to-Game
            & Brief, assets, reference video
            & \shortstack[l]{GDD, Godot project,\\feature demos}
            & \shortstack[l]{Mechanics, content, playability,\\design, VLM} \\
            
            GDD-to-Game
            & GDD, assets, reference video
            & \shortstack[l]{Godot project,\\feature demos}
            & \shortstack[l]{Mechanics, content,\\playability, VLM} \\
            
            Skeleton Completion
            & \shortstack[l]{Code skeleton, requirements,\\assets, reference video}
            & \shortstack[l]{Completed Godot project,\\feature demos}
            & \shortstack[l]{Mechanics, content, playability,\\scaffold, VLM} \\
            
            Bug Repair
            & \shortstack[l]{Faulty project, bug report,\\requirements}
            & Repaired Godot project
            & \shortstack[l]{Restoration, retained routes,\\preservation contracts, validity gates} \\
            
            Godot-to-Unity Porting
            & \shortstack[l]{Godot source, GDD, assets,\\reference video, Unity interface}
            & \shortstack[l]{Unity project,\\build instructions}
            & \shortstack[l]{Mechanics, playability, structure,\\visual quality, stability} \\
            
            \bottomrule
        \end{tabular}
    }
    \vspace{2pt}
    \raggedright \scriptsize 
    Feature demos are agent-written player-action sequences replayed by the evaluator. Content includes asset realization; Design assesses the authored GDD; Scaffold assesses completion and integration. VLM denotes rubric-based visual assessment. Build and interface checks are part of the porting evaluation.
\end{table}

\subsection{Game Instrumentation Interface}
\label{sec:interface}

Equivalent objects carry different names and scene hierarchies in
independently written games. A common instrumentation interface maps them to
semantic roles, standardized actions, and observable values. Godot submissions
declare scenes and endings in \texttt{gb\_levels.json}, tag objects with
groups such as \texttt{gb\_player} and \texttt{gb\_hazard}, and bind
task-relevant properties such as health, progress, and score; Unity
submissions provide equivalent bindings. The evaluator validates these
declarations, then instruments a working copy of the project with drivers and
probes that record positions, contacts, and state changes. A spike-contact
test, for instance, checks that touching a hazard is followed by the required
health change or death transition. Submissions supply only the bindings; the
checks and expected outcomes belong to the evaluator.

\subsection{Evaluation Protocol}
\label{sec:evaluation}

\paragraph{Functional evaluation.}
Requirements and reference executions define the expected structure, content,
and behavior, which evaluator-owned tests check through the instrumentation
interface. Construction evaluation uses two replay protocols. Certified routes replay input sequences authored and validated on the
reference game. Where specified, semantic start positions place the
player relative to relevant gameplay objects. The routes retain the
reference action timing and evaluate predefined goals and intermediate
milestones. These checks measure whether the candidate satisfies the
registered behavioral requirements under the prescribed input sequence.
A route earns no credit if its matched no-input control reaches the goal. Agent-authored demonstrations in
\texttt{demos.json} exercise features of the submitted implementation, each
from a fresh launch. A demonstrated outcome counts only if it does not also
occur in a no-input run of matched duration, and coverage is aggregated across
demonstrations without rewarding duplicates. The evaluator also records
gameplay during replay for visual assessment. Mechanics combines engine-state
checks with certified-route evidence; Playability combines demonstration
validity, feature coverage, and evidence that observed behavior depends on the
supplied actions.

Repair evaluation applies identical scenarios to the reference, faulty, and
repaired versions. A case qualifies only if the reference passes, the faulty
version fails the designated behaviors, and unaffected checks remain passing.
Repairs earn credit for restoration and preservation regardless of whether
their patches match the reference repair. Porting evaluation checks project
and interface readiness, the Unity build, and runtime behavior in the target
engine.

\paragraph{Visual evaluation.}
For the construction modes, a VLM scores frames from evaluator-recorded
gameplay against reference evidence and asset previews. The 41 game-specific rubrics, adapted from GameCraft-Bench \citep{luo2026gamecraft}, contain 658 items in four groups: Gameplay Realization, Content Composition, Guidance and Feedback, and Visual Presentation. Items define observable attainment criteria and score
ceilings for specified deficiencies, and each judgment cites frames and rubric
clauses while respecting design choices the task leaves open.

\paragraph{Score aggregation.}
For construction tasks, mechanics, content, and playability, together with
design for Brief-to-Game and scaffold for Skeleton Completion, form an
objective score normalized by its 85-point ceiling to $O\in[0,1]$; the component weights are listed in
Table~\ref{tab:score-weights} in
Appendix~\ref{app:score-aggregation}. Visual item scores in $[0,1]$ are
capped by any triggered deficiency ceiling, averaged within each group, and
combined with weights of 0.10, 0.18, 0.27, and 0.45, renormalized over
applicable groups, to give $V\in[0,1]$. The construction score is
\begin{equation}
    S=0.85O+0.15V.
    \label{eq:composite}
\end{equation}
A repair's score is the product of its restoration, retained-route,
preservation-contract, and validity-gate components, averaged across cases.
Porting uses a weighted aggregate of mechanics, playability, structure, visual
quality, and stability. All results are reported on a 0 to 100 scale.

\providecommand{\swegameneedspace}[1]{\par\begingroup\dimen0=\pagegoal\advance\dimen0 by-\pagetotal\ifdim\dimen0<#1\newpage\fi\endgroup}
\section{Experiments}
\label{sec:experiments}

Our experiments examine whether coding agents realize the intended gameplay
across the five tasks, how reference videos affect the resulting games, and how
closely automated evaluation agrees with human judgment. We also analyze
resource use and the implementation problems behind low scores.

\subsection{Experimental Setup}
\label{sec:experimental-setup}

We evaluate Qwen3.8 Flash\citep{qwen38omniflash}, Grok4.6\citep{xai2026grok46}, GPT-5.6 Luna\citep{openai2026gpt56}, Opus5\citep{anthropic2026opus5}, GLM5.3 Flash\citep{zai2026glm53}, and
Minimax M3\citep{minimax2026m3} on all 247 tasks, each with its own agent framework and tools, so
each reported configuration pairs a model with its development framework.
Table~\ref{tab:main-results} reports mean component and overall scores on a
0 to 100 scale under the protocol of Section~\ref{sec:evaluation}.
% Author: add exact model identifiers, framework versions, reasoning settings,
% execution budgets, available tools, and retry policies from the run manifests.

\subsection{Main Results}
\label{sec:main-results}

\paragraph{Performance across tasks.}
Opus5 achieves the highest overall score in all five tasks, from 50.38 in
brief-to-game generation to 83.46 in bug repair
(Table~\ref{tab:main-results}). Below it, the ranking depends on the task:
Grok4.6 is second in brief-to-game generation and GPT-5.6 Luna in the other
four. Every model scores higher on GDD-to-game implementation than on
brief-to-game generation, although the two tasks differ in both inputs and
scoring criteria.

\raggedbottom \begin{table}[t]
\centering
\caption{Component and overall scores on SWE-Game (0 to 100; higher is better).}
\label{tab:main-results}
\footnotesize
\setlength{\tabcolsep}{2.5pt}
\renewcommand{\arraystretch}{0.96}

% Brief-to-Game
\begin{tabular*}{\linewidth}{@{}p{0.15\linewidth}p{0.18\linewidth}@{\extracolsep{\fill}}cccccc@{}}
\toprule
Task & Model
& Mechanics
& Content
& Playability
& Design
& VLM
& \textbf{Total} \\
\cmidrule(l){2-8}
\multirow{6}{*}{\shortstack[l]{Brief-to-Game}}
& Qwen3.8 Flash & 21.10 & 23.32 & 28.38 & 65.85 & 40.40 & 27.00 \\
& Grok4.6      & 25.66 & 35.34 & 61.21 & 70.00 & 48.39 & 39.01 \\
& GPT-5.6 Luna & 24.17 & 21.29 & 55.51 & 92.68 & 60.05 & 34.18 \\
& Opus5        & 27.55 & 48.26 & 80.99 & 53.85 & 69.23 & 50.38 \\
& GLM5.3 Flash & 17.25 & 33.07 & 44.51 & 61.33 & 33.03 & 31.05 \\
& Minimax M3   & 25.30 & 27.20 & 42.65 & 67.50 & 32.09 & 30.47 \\
\end{tabular*}
\par\nointerlineskip

% GDD-to-Game
\begin{tabular*}{\linewidth}{@{}p{0.15\linewidth}p{0.18\linewidth}@{\extracolsep{\fill}}ccccc@{}}
\midrule
& Model
& Mechanics
& Content
& Playability
& VLM
& \textbf{Total} \\
\cmidrule(l){2-7}
\multirow{6}{*}{\shortstack[l]{GDD-to-Game}}
& Qwen3.8 Flash & 25.07 & 32.25 & 64.75 & 49.61 & 37.35 \\
& Grok4.6      & 28.66 & 53.58 & 59.60 & 60.20 & 48.55 \\
& GPT-5.6 Luna & 31.52 & 53.52 & 76.57 & 67.06 & 52.67 \\
& Opus5        & 32.96 & 61.07 & 91.94 & 75.03 & 59.68 \\
& GLM5.3 Flash & 26.21 & 37.83 & 69.45 & 45.84 & 40.18 \\
& Minimax M3   & 32.84 & 38.05 & 72.51 & 46.19 & 42.58 \\
\end{tabular*}
\par\nointerlineskip

% Skeleton Completion
\begin{tabular*}{\linewidth}{@{}p{0.15\linewidth}p{0.18\linewidth}@{\extracolsep{\fill}}cccccc@{}}
\midrule
& Model
& Mechanics
& Content
& Playability
& Scaffold
& VLM
& \textbf{Total} \\
\cmidrule(l){2-8}
\multirow{6}{*}{\shortstack[l]{Skeleton\\Completion}}
& Qwen3.8 Flash & 24.60 & 20.99 & 65.51 & 91.15 & 58.40 & 39.19 \\
& Grok4.6      & 26.92 & 16.03 & 80.75 & 86.36 & 70.08 & 40.76 \\
& GPT-5.6 Luna & 29.57 & 17.70 & 82.35 & 88.44 & 73.87 & 43.10 \\
& Opus5        & 36.22 & 36.93 & 93.77 & 94.62 & 78.36 & 54.06 \\
& GLM5.3 Flash & 22.63 & 11.97 & 82.70 & 79.13 & 49.90 & 34.38 \\
& Minimax M3   & 21.87 & 20.03 & 79.09 & 74.39 & 47.49 & 35.69 \\
\end{tabular*}
\par\nointerlineskip

\begin{tabular*}{\linewidth}{@{}p{0.15\linewidth}p{0.18\linewidth}@{\extracolsep{\fill}}ccccc@{}}
\midrule
& Model
& Restoration
& \shortstack{Retained\\routes}
& \shortstack{Preservation\\contracts}
& \shortstack{Validity\\gates}
& \textbf{Total} \\
\cmidrule(l){2-7}
\multirow{6}{*}{\shortstack[l]{Bug Repair}}
& Qwen3.8 Flash & 44.40 & 98.61 & 80.56 & 98.61 & 26.13 \\
& Grok4.6      & 56.52 & 96.88 & 85.94 & 99.72 & 52.44 \\
& GPT-5.6 Luna & 69.62 & 98.35 & 86.08 & 99.72 & 58.05 \\
& Opus5        & 90.18 & 99.51 & 93.17 & 99.72 & 83.46 \\
& GLM5.3 Flash & 44.34 & 96.00 & 75.51 & 99.85 & 27.55 \\
& Minimax M3   & 35.27 & 96.05 & 76.32 & 94.74 & 27.36 \\
\end{tabular*}
\par\nointerlineskip

\begin{tabular*}{\linewidth}{@{}p{0.15\linewidth}p{0.18\linewidth}@{\extracolsep{\fill}}cccccc@{}}
\midrule
& Model
& Mechanics
& Playability
& Structure
& Visual
& Stability
& \textbf{Total} \\
\cmidrule(l){2-8}
\multirow{6}{*}{\shortstack[l]{Godot-to-Unity\\Porting}}
& Qwen3.8 Flash & 57.31 & 50.39 & 50.27 & 36.78 & 55.20 & 51.23 \\
& Grok4.6      & 61.71 & 50.26 & 78.76 & 35.40 & 77.50 & 59.04 \\
& GPT-5.6 Luna & 63.26 & 51.44 & 82.87 & 31.60 & 76.80 & 59.85 \\
& Opus5        & 70.66 & 65.88 & 90.53 & 58.40 & 88.60 & 72.40 \\
& GLM5.3 Flash & 50.37 & 38.32 & 63.13 & 12.13 & 58.30 & 44.33 \\
& Minimax M3   & 53.44 & 48.31 & 42.74 & 49.28 & 50.32 & 49.62 \\
\bottomrule
\end{tabular*}

\vspace{0.6em}
\begin{minipage}{\linewidth}
\footnotesize
Content includes asset realization; Design assesses the authored GDD; Scaffold assesses completion and integration. For construction tasks, Mechanics includes certified routes that replay reference inputs, using semantic start positions where specified. Playability combines demonstration validity, feature coverage, and input-dependent behavior assessed against matched no-input controls. Totals follow Section~\ref{sec:evaluation}; Bug Repair totals average per-task products of the component factors.
\end{minipage}
\end{table}
\paragraph{Performance by component.}
In skeleton completion, Content scores range from 11.97 to 36.93,
indicating limited realization of the required content across models.
In bug repair, Restoration varies more widely across models
(35.27--90.18) than Retained routes (96.00--99.51) or Preservation
contracts (75.51--93.17). Both restoration and preservation contribute
to the repair score.
In porting, GPT-5.6 Luna and Opus5 score higher in Structure than in
Playability and Visual. GPT-5.6 Luna obtains 82.87, 51.44, and 31.60
in these components, respectively, while Opus5 obtains 90.53, 65.88,
and 58.40.

\subsection{Resource Use and Error Analysis}
\label{sec:agent-analysis}
\label{sec:failure-analysis}

\begin{figure}[t]
    \centering
    \includegraphics[width=\linewidth]{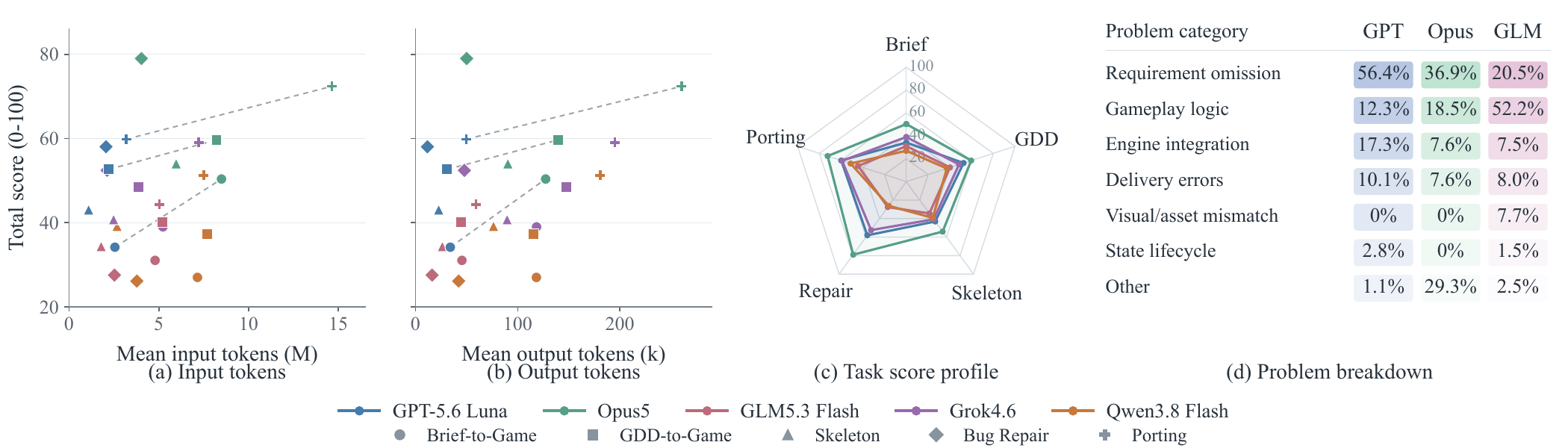}
    \caption{Resource use, task performance, and problem composition on SWE-Game. Panels (a) and (b) compare task scores against mean input and output tokens, respectively; colors identify models and shapes identify tasks. Dashed lines connect the selected within-task nondominated configurations. Panel (c) summarizes total scores across the five tasks. Panel (d) shows the distribution of problem categories within each model.}
    \label{fig:pareto-draft}
\end{figure}

\paragraph{Performance and resource use.}
Figure~\ref{fig:pareto-draft}(a) and (b) relate scores to mean cumulative
input and output tokens, respectively, over the recorded runs, following
the resource analyses of Terminal-Bench and
GameCraft-Bench~\citep{merrill2026terminal,luo2026gamecraft}; accounting
rules are given in Appendix~\ref{app:resource-accounting}.
Panel (c) summarizes total scores across the five tasks.
The resource statistics characterize model and framework configurations,
whose token accounting and tools differ. A configuration is nondominated
when no other configuration in the same task achieves at least its score
with no more of the plotted resource and is strictly better in one of
the two. In brief-to-game generation, for example, Opus5 attains the
highest score with 8.49M input tokens, 127.34k output tokens, and 111.37
tool calls, compared with 2.56M, 34.05k, and 57.15, respectively, for
GPT-5.6 Luna.

\par\medskip
\noindent
\textbf{Implementation error patterns.}\quad
Figure~\ref{fig:pareto-draft}(d) summarizes the types of implementation
problems observed for each model. Requirement omissions are the most
common category for GPT and Opus, accounting for 56.4\% and 36.9\% of
their categorized problems, respectively. Engine-integration errors
account for another 17.3\% of GPT's problems. For GLM, gameplay logic
errors are the largest category, at 52.2\%. Percentages are computed
over categorized problems within each model, with a single game
potentially contributing multiple problems.

Requirement omissions and gameplay-logic errors call for different forms of validation. Tracking each task obligation against the implementation and its demonstrations targets missing behavior. Exercising trigger conditions and inspecting state changes targets behavior that is present but incorrectly implemented. Engine-integration errors add a further need to check connections among scripts, scenes, and resources. These categories motivate test plans that combine requirement coverage with interaction and integration checks, providing concrete targets for future improvements to agent workflows.

\par
\flushbottom \subsection{Reference-Video Ablation} \label{sec:ablations} \begin{table}[t] \centering \begin{minipage}{0.64\linewidth} \caption{Reference-video ablation in brief-to-game generation.
Scores are on a 0 to 100 scale.}
\label{tab:ablations}
\footnotesize
\setlength{\tabcolsep}{2pt}
\renewcommand{\arraystretch}{1.06}
\begin{tabular*}{\linewidth}{@{\extracolsep{\fill}}lrrrrr@{}}
\toprule
Video & \shortstack{Game-\\play} & Content & Feedback & Visual
& \shortstack{\textbf{VLM}\\\textbf{Total}} \\
\midrule
\multicolumn{6}{@{}l}{\textbf{GPT-5.6 Luna}} \\
On  & 68.87 & 60.69 & 63.24 & 55.92 & \textbf{60.05} \\
Off & 58.57 & 66.70 & 53.10 & 49.44 & 54.45 \\
\midrule
\multicolumn{6}{@{}l}{\textbf{Opus5}} \\
On  & 73.54 & 70.55 & 78.06 & 62.45 & \textbf{69.23} \\
Off & 77.48 & 80.48 & 76.73 & 53.45 & 67.00 \\
\bottomrule
\end{tabular*} \end{minipage} \end{table}

Table~\ref{tab:ablations} compares brief-to-game generation with and without
the reference video for GPT-5.6 Luna and Opus5. Only the agent's input
changes; the evaluator receives the same reference evidence for each game. The
VLM total is higher with video for both models, by 5.60 points for GPT-5.6 Luna
and 2.23 for Opus5, and Visual Presentation, the most heavily weighted group,
rises by 6.48 and 9.00. The other components move in mixed directions. With
video, GPT-5.6 Luna gains 10.30 in Gameplay Realization and 10.14 in Guidance
and Feedback, while Opus5 changes by $-3.94$ and $+1.33$; Content Composition
is lower with video for both models, by 6.01 and 9.93. The consistent gain is
therefore in presentation, consistent with gameplay videos conveying the target
appearance. Both conditions evaluate the same complete set of 41 games. The reported differences therefore compare video-on and video-off performance on an identical task set.

\swegameneedspace{8\baselineskip}
\subsection{Evaluation Validity}
\label{sec:evaluation-validity}

\begin{wraptable}{r}{0.50\textwidth}
\vspace{-13pt}
\centering
\caption{Human validation of executable and visual evaluation.}
\label{tab:evaluation-validity}
\footnotesize
\setlength{\tabcolsep}{2pt}
\renewcommand{\arraystretch}{1.04}
\begin{tabular*}{\linewidth}{@{\extracolsep{\fill}}lrr@{}}
\toprule
\multicolumn{3}{@{}l}{\textbf{Functional judgments (\%)}} \\
Metric & Executable & Video VLM \\
\midrule
Correct acceptance & 90.93 & 81.41 \\
Defect detection & 94.25 & 75.40 \\
Balanced accuracy & 92.59 & $\approx 78.41$ \\
Reviewed games/samples & 100/1200 & 100/1200 \\
\bottomrule
\end{tabular*}
\par\vspace{4pt}
\begin{tabular*}{\linewidth}{@{\extracolsep{\fill}}lr@{}}
\toprule
\multicolumn{2}{@{}l}{\textbf{Visual judgments}} \\
Metric & Rubric VLM \\
\midrule
Spearman $\rho$ & 0.829 \\
Mean absolute error & 0.103 \\
Repeated-score SD & 0.034 \\
Reviewed clips & 200 \\
\bottomrule
\end{tabular*}
\par\vspace{2pt}
\raggedright
Balanced accuracy is the unweighted mean of correct acceptance
and defect detection rates. Visual error and SD use a [0,1] scale.
\vspace{-5pt}
\end{wraptable}
Table~\ref{tab:evaluation-validity} compares automated judgments with human
assessments. Functional validation uses a stratified sample of 1,200 human-labeled
assertions from 100 games under Modes 1--3, with nine Mechanics and
three Content assertions per game. These projects exclude the reference
and injected-fault cases used to construct the evaluator. Executable
checks correctly accept 411/452 correct assertions and detect 705/748
defective ones, achieving 92.59\% balanced accuracy versus approximately
78.41\% for the video-based VLM judge, with an 18.85-percentage-point
advantage in defect detection. For visual quality, rubric-based VLM scores on 200 clips correlate with human ratings at Spearman $\rho = 0.829$, with a mean absolute error of 0.103 on a $[0,1]$ scale and a within-clip standard deviation of 0.034 across repeated judgments. These results support assigning behavioral correctness to executable checks and perceptual quality to the visual rubric.

\section{Conclusion}
\label{sec:conclusion}

We introduced \bench{}, a reference-grounded benchmark covering game construction, completion, repair, and cross-engine porting. Executable reference games provide concrete targets, while a shared instrumentation interface connects gameplay requirements to runtime evidence. Combined with visual assessment, this framework evaluates both behavioral correctness and perceptual quality. Results across six models reveal a persistent gap between producing playable projects and faithfully implementing the intended mechanics and content. \bench{} provides a basis for studying coding agents across the development and maintenance of interactive software.

\section*{Limitations and Future Work}

The current benchmark contains 41 reference games, and its coverage of large projects and long-term development remains limited. Each model is evaluated with a single agent framework, so the results reflect their combined capabilities. Godot-to-Unity porting remains an initial study of engine migration. Future work will expand the game collection, compare frameworks under controlled budgets, and evaluate longer development workflows. Broader human assessment will further refine the visual rubrics and clarify how automated scores relate to player experience.
\subsection*{AI use statement}

In this work, we used generative AI tools to assist with literature organization and scientific writing. All AI-assisted code were reviewed by the authors.We did not use generative AI to fabricate experimental results or replace human judgments of game validity. We take full responsibility for the final content of this work, including all text, claims, code, data, and other artifacts produced with the assistance of generative AI.

\subsection*{Ethics}
This work does not collect or release personal, sensitive, or personally identifiable data. The game projects and assets were collected from public sources, and their provenance and licenses were recorded during dataset construction. Generated projects are executed in isolated evaluation environments without access to external services, credentials, or private user data. Because the benchmark may enable the reconstruction of games resembling existing works, users should respect copyright, licensing, and attribution equirements when creating or releasing game content.

\bibliography{iclr2027_conference}
\bibliographystyle{iclr2027_conference}

\appendix
\ifcsname sweappendixpage\endcsname\else

\newsavebox{\sweappendixpagebox}
\newdimen\sweappendixpageheight

\newenvironment{sweappendixpage}{%
  \clearpage
  \begin{lrbox}{\sweappendixpagebox}%
  \begin{minipage}[t]{\textwidth}%
  \small
  \setlength{\parskip}{3pt}%
  \setlength{\abovecaptionskip}{4pt}%
  \setlength{\belowcaptionskip}{4pt}%
  \setlength{\abovedisplayskip}{5pt}%
  \setlength{\belowdisplayskip}{5pt}%
  \tcbset{unbreakable,before skip=4pt,after skip=4pt}%
}{%
  \par
  \end{minipage}%
  \end{lrbox}%
  \sweappendixpageheight=
    \dimexpr\ht\sweappendixpagebox+\dp\sweappendixpagebox\relax
  \noindent
  \makebox[\textwidth][c]{%
    \ifdim\sweappendixpageheight>0.96\textheight
      \resizebox*{!}{0.96\textheight}{%
        \usebox{\sweappendixpagebox}%
      }%
    \else
      \usebox{\sweappendixpagebox}%
    \fi
  }%
  \par
}

\fi

\clearpage
\section{Reference Game Composition}
\label{app:reference-games}

Table~\ref{tab:reference-game-catalog} summarizes the 41 reference games used in SWE-Game. The collection consists of 24 2D and 17 3D games and covers projects with varying gameplay structures and scales, from compact single-mechanic games to multi-stage games with interacting mechanics and persistent progression.
In aggregate, the reference games contain 168,335 lines of gameplay GDScript, 1,587 scenes, 22,140 scene-tree nodes, and 138 gameplay recordings. Gameplay-code size ranges from 1,153 to 11,024 lines per project, with a median of 3,367 lines, while the number of scenes ranges from 7 to 108, with a median of 32.

\begingroup
\centering
\captionof{table}{\textbf{Composition of the 41 reference games in \bench{}.}
}
\label{tab:reference-game-catalog}

\scriptsize
\setlength{\tabcolsep}{3.5pt}
\renewcommand{\arraystretch}{1.0}

\begin{tabularx}{\textwidth}{
    @{}
    >{\raggedright\arraybackslash}p{0.170\textwidth}
    >{\centering\arraybackslash}p{0.035\textwidth}
    >{\raggedright\arraybackslash}p{0.145\textwidth}
    >{\raggedright\arraybackslash}X
    >{\raggedleft\arraybackslash}p{0.050\textwidth}
    >{\raggedleft\arraybackslash}p{0.040\textwidth}
    @{}
}
\toprule
\textbf{Game}
& \textbf{Dim}
& \textbf{Category}
& \textbf{Core gameplay}
& \textbf{Code}
& \textbf{Scenes} \\
\midrule

\multicolumn{6}{@{}l}{\textbf{2D Games (24)}} \\
\midrule

Arc Wing
& 2D & Shooting
& Survive staged enemy waves in a vertical shooter.
& 3,299 & 25 \\

Ballast Yard
& 2D & Puzzle
& Route objects through a deterministic grid simulation.
& 2,588 & 33 \\

Last Cat on the Wall
& 2D & Strategy \& defense
& Balance defensive economy, shooting, and traversal.
& 4,431 & 31 \\

Don't Stop
& 2D & Shooting
& Clear a compact top-down science-fiction combat course.
& 5,780 & 108 \\

The Ember Key
& 2D & Action \& adventure
& Fight and navigate through a top-down dungeon escape.
& 2,672 & 39 \\

Ember \& Tide
& 2D & Puzzle
& Coordinate two elemental characters to reach paired exits.
& 3,981 & 39 \\

\resizebox{\linewidth}{!}{Another Gentleman's Adventure}
& 2D & Shooting
& Combine side-scrolling traversal with ranged combat.
& 2,703 & 32 \\

Gunfire Dungeon
& 2D & Shooting
& Clear rooms, purchase weapons, and defeat a boss.
& 3,029 & 42 \\

Harvest Ledger
& 2D & Simulation \& management
& Farm, trade, and schedule resources under deadlines.
& 6,697 & 86 \\

Hazard Circuit
& 2D & Platform \& parkour
& Complete deterministic obstacle and hazard courses.
& 5,521 & 35 \\

Hive Flight
& 2D & Platform \& parkour
& Avoid obstacles during a side-scrolling chase.
& 3,035 & 87 \\

Lantern Vigil
& 2D & Action \& adventure
& Complete a single-run top-down ninja combat challenge.
& 2,770 & 21 \\

DYNAMO
& 2D & Puzzle
& Manipulate timed world states and player-created terrain.
& 2,829 & 28 \\

Pixel Platformer
& 2D & Platform \& parkour
& Complete four themed precision-platforming rooms.
& 4,313 & 7 \\

Pulse Lane
& 2D & Rhythm
& Hit notes across four vertically scrolling lanes.
& 2,820 & 25 \\

Relic Runner
& 2D & Action \& adventure
& Clear rooms, transport a relic, and avoid damage.
& 2,807 & 30 \\

Shadow Walker
& 2D & Stealth
& Navigate guarded spaces as an unarmed infiltrator.
& 2,696 & 21 \\

Soccer Course
& 2D & Sports
& Tackle, pass, shoot, and complete a tournament.
& 2,741 & 18 \\

Super Dungeon Delve
& 2D & Action \& adventure
& Fight through four dungeon floors while collecting loot.
& 1,153 & 20 \\

\resizebox{\linewidth}{!}{Bannerfall: Tiny Kingdoms}
& 2D & Strategy \& defense
& Command a hero-led army in strategic battles.
& 5,926 & 30 \\

Vaultline
& 2D & Platform \& parkour
& Traverse momentum courses with a shared death budget.
& 4,965 & 38 \\

Volley Break
& 2D & Sports
& Play a paddle duel across a destructible block wall.
& 2,475 & 34 \\

Wizard Chase
& 2D & Maze chase
& Collect treasure across five increasingly hostile mazes.
& 3,109 & 30 \\

Sprout Market
& 2D & Simulation \& management
& Produce, transport, and sell goods in real time.
& 3,915 & 32 \\

\midrule
\multicolumn{6}{@{}l}{\textbf{3D Games (17)}} \\
\midrule

Beacon Relay
& 3D & Platform \& parkour
& Traverse moving hazards and deliver power cells.
& 4,146 & 33 \\

Bush 522
& 3D & Driving \& flight
& Take off, pass checkpoints, manage flight, and land.
& 1,408 & 17 \\

Canopy Dash
& 3D & Platform \& parkour
& Switch lanes, jump, slide, and turn around obstacles.
& 6,992 & 67 \\

Manifest Run
& 3D & Driving \& flight
& Route multiple deliveries under overlapping deadlines.
& 3,367 & 78 \\

CityCab Rush
& 3D & Driving \& flight
& Transport passengers before the fare timer expires.
& 4,605 & 17 \\

Neon Lockdown
& 3D & Shooting
& Infiltrate a hostile facility and survive its combat run.
& 3,099 & 40 \\

Deepgrid
& 3D & Shooting
& Explore six vault floors and defeat the overseer.
& 5,110 & 40 \\

DOGWALK
& 3D & Action \& adventure
& Complete errands with an autonomous leashed companion.
& 11,024 & 38 \\

Hurry Curry!
& 3D & Simulation \& management
& Plan and prepare dishes under time constraints.
& 5,108 & 76 \\

Kindle Relay
& 3D & Action \& adventure
& Explore a spherical world and relight six beacons.
& 3,585 & 24 \\

Pixel Sabotage
& 3D & Stealth
& Infiltrate a guarded environment without detection.
& 1,795 & 79 \\

Apex Circuit
& 3D & Driving \& flight
& Complete a three-lap race against four opponents.
& 3,875 & 67 \\

Sands of the Restless
& 3D & Shooting
& Survive enemy waves and purchase combat upgrades.
& 9,720 & 15 \\

Terraforge
& 3D & Survival \& crafting
& Modify terrain and complete a seeded crafting objective.
& 8,165 & 26 \\

Scrapline Siege
& 3D & Shooting
& Defend an arena against successive enemy waves.
& 5,517 & 35 \\

Twin Holds
& 3D & Strategy \& defense
& Compete against a scripted opponent on a hexagonal board.
& 2,983 & 17 \\

\mbox{Where The Dead Lie}
& 3D & Horror
& Explore a hostile environment and search for resources.
& 1,581 & 27 \\

\bottomrule
\end{tabularx}

\par\endgroup
\section{Task Construction and Development Modes}
\label{app:tasks}

Each reference game supports five development settings by varying the
specification and the starting implementation. Modes~1--3 provide assets and
reference gameplay video, but differ in whether the starting point is a
design brief, a reviewed GDD, or a supplied integration scaffold. Mode~4
provides a faulty implementation and a symptom report. Mode~5 provides the
Godot implementation and a Unity target project. The following descriptions
specify the task artifacts and deliverables.

\subsection{Brief-to-Game}
\label{app:brief-to-game}

Mode~1 starts from a project-specific design target, an asset palette, and
reference gameplay video. The generator reads a curated brief catalog and
renders its \texttt{title}, \texttt{experience}, \texttt{scope}, and
\texttt{freedom} fields into \texttt{statement.md}. Where applicable, it adds
the task's required extended-control obligations. This construction provides
a concise target experience and scope while leaving concrete rules and
implementation choices to the developer; it does not automatically summarize
or distribute the reference GDD.

For example, the catalog entry for \emph{Beacon Relay} describes a compact
third-person traversal course with spatial progression, recoverable falls,
and an explicit completion state. It leaves course topology, movement
tuning, beacon rules, checkpoints, and scenic composition open. The box below
summarizes this actual entry rather than presenting an agent instruction.

\begin{taskbox}[breakable,title={Mode 1 Example: Beacon Relay task specification}]
Input artifacts:
  statement.md    Project-specific design target
  assets/         Supplied media palette
  video/          Reference gameplay
  interface/      Godot submission contract

Catalog entry (abridged):
  experience: Traverse a 3D course and complete a beacon relay.
  scope:      Compact third-person platformer; spatial
              progression, recoverable failure, and completion.
  freedom:    Course topology, movement tuning, beacon rule,
              checkpoints, and scenic composition.

Deliverables:
  GDD.md + runnable Godot project + demos.json
  (legacy whole-run ops.json is also accepted)
\end{taskbox}

\begin{taskbox}[breakable,title={Mode 1 Example: Beacon Relay generated statement.md}]
# Beacon Relay --- task brief

This is a design target, not an implementation specification. Build your own GDD, scene structure, rules, and code.

## Target player experience

Cross a readable 3D traversal course, use movement to activate or reach a sequence of beacons, and finish the relay.

## Scope boundary

A compact third-person 3D platformer with spatial progression, recoverable falls or failure, and an explicit final completion state.

## Creative freedom

You may choose the course topology, movement tuning, beacon rule, checkpoints, and scenic composition.
\end{taskbox}

\subsection{GDD-to-Game}
\label{app:gdd-to-game}

Mode~2 provides a reviewed task GDD, assets, reference gameplay video, and the
Godot submission interface. The generator copies the reviewed per-game
document to \texttt{GDD.md}. It also writes \texttt{DEMONSTRATIONS.md}, listing
the observable behavior claims associated with the task's mechanic checks.
Executable predicates and reference solutions remain evaluator-only.

\begin{taskbox}[breakable,title={Mode 2 Example: Beacon Relay GDD}]
Input artifacts:
  GDD.md             Reviewed task design
  DEMONSTRATIONS.md  Observable feature obligations
  assets/            Supplied media palette
  video/             Reference gameplay
  interface/         Godot submission contract

Starting implementation:
  No reference gameplay source is supplied.

Deliverables:
  Runnable Godot project + demos.json
  (legacy whole-run ops.json is also accepted)

# Beacon Relay -- Task GDD

## Target experience

Carry unstable energy cells across floating meadow islands and deposit them
before their fuse expires. Use movement devices and well-timed stomps to turn
threats into additional delivery time.

## Design pillars

1. The fuse is the clock: picking up a cell creates immediate visible pressure.
2. The return trip is the challenge: geometry and devices affect how the cell
   is carried back to the beacon.
3. Threats can be resources: a stomp while carrying refunds time, while ordinary
   contact costs shields.
4. Fixed, readable motion: camera yaw uses eight 45-degree steps and required
   geometry remains inside the measured movement envelope.

## Scope

- Three 3D levels and eleven total cells.
- Three hostile roles and seven device families.
- One heart per level and three shields.
- A complete run lasts approximately 6--9 minutes at fixed 60 Hz.
- Gameplay is deterministic and contains no procedural randomness.

## Controls

Move:       gb_up, gb_down, gb_left, gb_right
Jump:       gb_jump
Dive:       gb_action
Pause:      gb_pause
Restart:    gb_reset

The player moves relative to the camera, jumps from the ground or once in the
air, and uses gb_action while airborne to perform a fast-fall stomp.

## Core mechanics

Cell fuse:
  Touching a pedestal cell while not carrying another cell starts a
  level-dependent countdown of 13, 12, or 11 seconds. The cell changes color
  from green to amber to red as the fuse expires.

Deposit:
  When a player carrying a cell overlaps a beacon, relay progress increases,
  one beacon segment lights, and the fuse stops.

Shields:
  Hostile or hazard contact and falls remove one of three shield pips.
  Reaching zero shields produces failure.

Stomp:
  Contacting a hostile from above while falling defeats or damages it, bounces
  the player, refreshes the air jump, and refunds 2.5 seconds while carrying a
  cell. Side or lower contact costs a shield.

## Progression

Meadow Relay:
  Teach cell delivery, fuse pressure, moving and crumble platforms, spikes, and
  the first stomp-refund opportunity. Deposit three cells.

Saw Gully:
  Vary the route with four cells, springs, saw timing, and conveyors.
  Deposit four cells.

Frostline Summit:
  Combine the mechanics with four cells, a timed gate, a large moving lift,
  tighter fuse timing, hazards, and stomp opportunities. Deposit all cells to
  reach full victory.

## End states

Level clear occurs when every cell in the current level is physically deposited
before the level clock expires. Full victory occurs after all cells in the third
level are deposited. Failure occurs when shields reach zero or the level clock
expires before the delivery quota is met. Nonterminal hits and falls respawn the
player at the last checkpoint while preserving delivered-cell progress.

## Interface and acceptance

The player belongs to gb_player; hostiles, hazards, collectibles, interactive
devices, checkpoints, doors, and goals use their corresponding gb_* semantic
groups. The project declares three levels and distinct victory and defeat
endings, and exposes live progress, health, and timer values through
gb_levels.json.

All levels, devices, pause, restart, failure, and victory transitions must run
without fatal errors. The full ordinary-input route must be completable, fuse
expiry must return a cell to its pedestal, and the visible presentation must
communicate cell urgency, beacon progress, hostile facing, device state, shield
loss, and the current goal.
\end{taskbox}

\subsection{Skeleton Completion}
\label{app:skeleton-completion}

Mode~3 supplies a newly generated minimal Godot scaffold. The implementation
constructs this scaffold from public task obligations, rather than removing
gameplay code from the reference project. It contains an entry scene,
addressable placeholder level and ending scenes, InputMap bindings,
\texttt{gb\_levels.json}, and a player adapter in the
\texttt{gb\_player} group. The adapter reads the required actions and axes;
declared numeric slots initially expose placeholder values.

\begin{taskbox}[breakable,title={Mode 3 Example: Beacon Relay skeleton task}]
Input artifacts:
  SKELETON_TASK.md
  DEMONSTRATIONS.md
  assets/ + video/ + interface/

Generated scaffold:
  skeleton/project.godot
  skeleton/gb_levels.json
  skeleton/gb_player_adapter.gd
  skeleton/levels/level_01.tscn
  skeleton/levels/level_02.tscn
  skeleton/levels/level_03.tscn
  skeleton/endings/victory.tscn
  skeleton/endings/defeat.tscn

Task:
  Complete the minimal GB scaffold. This is a new implementation-neutral
  scaffold, not a redacted copy of the corpus game's internal architecture.
  It fixes only the public integration surface: an importable project entry,
  addressable placeholder levels and endings, InputMap actions, and one
  gb_player adapter.

Implementation freedom:
  Add, delete, or reorganize all other gameplay files and use any 2D or 3D
  architecture. Keep the supplied entry, level, and ending paths importable.
  Keep gb_player_adapter.gd attached to GBPlayerAdapter in each supplied level.
  The adapter may be rewritten internally.

Public GB obligations:
  Minimum addressable level scene paths: 3
  Extended actions: none
  Analog axes: none
  Required semantic groups:
    gb_player, gb_enemy, gb_hazard, gb_collectible,
    gb_interactive, gb_checkpoint, gb_door, gb_goal
  Required numeric slots:
    progress, health, timer
  Distinct failure ending required: true

Required observable behaviors:
  - Picking up a live relay cell removes it from the world collectible census.
  - A successful stomp can remove a defeatable hostile.
  - Depositing a carried cell advances relay progress.

Implementation flow:
  1. Import and launch the scaffold before editing.
  2. Read gb_player_adapter.gd and route its public action and state bindings
     into the implemented player and game state.
  3. Replace the placeholder level and ending contents with playable gameplay.
  4. Drive every required numeric property from live state.
  5. Preserve the declared InputMap identifiers and manifest paths.
  6. Verify cold launch, ordinary play, failure/restart, full completion, and
     a matched no-input negative control.

Deliverables:
  Completed Godot project + demos.json
  (legacy full-clear ops.json is also accepted)
  No GDD is required in this mode.
\end{taskbox}
\subsection{Bug Repair}
\label{app:bug-repair}

Mode~4 provides an existing Godot project with an injected gameplay defect,
a player-facing report, and gameplay requirements. For a registered case,
the generator creates a sanitized reference copy, applies the selected fault
bundle, and validates the reference--mutant behavior difference before
packaging the faulty project. Case selection uses a stable
\texttt{case\_id}, allowing multiple repair tasks to be derived from one game.

\begin{taskbox}[breakable,title={Mode 4: repair task artifacts}]
Input artifacts:
  game/                     Runnable faulty Godot project
  BUG_REPORT.md             Player-visible symptom report
  GAMEPLAY_REQUIREMENTS.md  Product behavior requirements
  BUGFIX_TASK.md            Maintenance task contract
  interface/                Godot submission contract

Starting implementation:
  Selected causal edits applied to a sanitized reference.
  A case may span multiple files or subsystems.

Deliverable:
  Repaired Godot project preserving unaffected behavior.
\end{taskbox}

\subsection{Godot-to-Unity Porting}
\label{app:godot-to-unity}

Mode~5 provides sanitized Godot source, the reviewed GDD, assets, reference
gameplay video, a prebuilt Unity target workspace, and the Unity interface
contract. The source implementation exposes the game's logic and resource
organization, while the GDD and video describe the behavior and presentation
to preserve. The benchmark uses Godot~4.5.1 and targets Unity
Editor~\texttt{6000.3.23f1}.

\begin{taskbox}[breakable,title={Mode 5: source and target artifacts}]
Input artifacts:
  source_godot/     Sanitized Godot implementation
  GDD.md            Reviewed task design
  assets/ + video/  Media and gameplay reference
  target_unity/     Supplied Unity project
  unity_interface/ Public port and observation contracts

Target project:
  Assets/Game/                         Candidate gameplay
  Assets/GameBenchmark/gb_interface.json
  Assets/GameBenchmarkSDK/             Locked SDK
  Packages/                            Locked dependencies
  ProjectSettings/ProjectVersion.txt   Locked editor version

Deliverables:
  Unity project + ops.json + BUILD.md
\end{taskbox}

Gameplay code belongs under \texttt{Assets/Game/}. The supplied SDK, package
manifests, lockfile, and editor-version file are protected by integrity
checks. The seeded interface declares scene addresses, action channels,
semantic roles, numeric slots, and outcome capabilities. The entry scene may
be a declared gameplay scene or a separate bootstrap scene. The required
input channels must operate the port's gameplay, and each declared level must
expose the specified player marker and task bindings.

\texttt{BUILD.md} documents a reproducible Linux Player build.
\texttt{ops.json} supplies an input witness for the port. Assets must be
imported into the Unity implementation; bundling or launching the Godot
runtime is excluded by the port contract. The target dependency set includes
a locked local Input System package. The task permits different component and
scene architectures while requiring the intended controls, mechanics,
progression, content, and visible feedback to be preserved.
\clearpage
\section{Game Instrumentation Interface}
\label{app:interface}

The interface specifies semantic entities, input actions, state bindings,
and scene/outcome declarations. Godot tasks share
\texttt{contract.json}; Unity porting tasks use a separate engine-specific
contract. Table~\ref{tab:interface-spec} summarizes their bindings.
The Godot manifest \texttt{gb\_levels.json} is one part of the interface,
not its complete specification.

\begingroup
\centering
\captionof{table}{Instrumentation interface and engine-specific bindings.}
\label{tab:interface-spec}
\small
\setlength{\tabcolsep}{4pt}
\renewcommand{\arraystretch}{1.08}
\begin{tabularx}{\linewidth}{@{}
    >{\raggedright\arraybackslash}p{0.17\linewidth}
    >{\raggedright\arraybackslash}X
    >{\raggedright\arraybackslash}X
@{}}
\toprule
Element & Godot & Unity \\
\midrule

Entities &
Scene-tree groups: required \texttt{gb\_player}; optional
\texttt{gb\_goal}, \texttt{gb\_enemy}, \texttt{gb\_hazard},
\texttt{gb\_collectible}, \texttt{gb\_checkpoint},
\texttt{gb\_door}, \texttt{gb\_interactive}. &
\texttt{GBEntity} components with semantic roles and stable IDs. \\

Actions &
InputMap bindings read by gameplay code; canonical actions,
optional extended actions and analog axes. &
Manifest fields \texttt{supported\_actions},
\texttt{required\_actions}, and \texttt{analog\_axes},
backed by gameplay input handling. \\

State &
\texttt{numeric} maps standard slots to bare property names;
\texttt{device\_ids} identifies individual objects. &
\texttt{GBTelemetry} and entity-level
\texttt{GBObservableState}. \\

Scenes &
\texttt{levels} lists playable scene paths;
\texttt{endings} maps outcome names to scenes. &
\texttt{entry\_scene} and \texttt{levels} declare scenes;
\texttt{GBOutcome} reports outcome events. \\

Manifest &
\texttt{gb\_levels.json}, validated against
\texttt{gb\_interface.schema.json}. &
\texttt{gb\_interface.json}, following the Unity port contract
(schema version~3). \\

Observations &
Engine-owned geometry, overlaps, scene state, object state,
and resource observations. &
Entity geometry, contacts, lifecycle, numerical state,
scene lifecycle, and outcome events. \\

\bottomrule
\end{tabularx}
\par\endgroup

\begin{taskbox}[breakable,title={Godot interface contract: registered schema excerpt}]
{
  "groups": {
    "required": ["gb_player"],
    "optional": [
      "gb_collectible", "gb_goal", "gb_hazard", "gb_enemy",
      "gb_checkpoint", "gb_door", "gb_interactive"
    ]
  },
  "actions": [
    "gb_left", "gb_right", "gb_up", "gb_down",
    "gb_jump", "gb_action", "gb_pause", "gb_reset"
  ],
  "numeric_slots": ["progress", "health", "timer", "score"],
  "manifest": {
    "required": ["levels", "endings"],
    "optional": [
      "numeric", "device_ids", "anchor_camera", "audio_buses",
      "level_clear", "level_entry",
      "extended_actions", "analog_axes"
    ]
  }
}
\end{taskbox}
\section{Mode-Specific Evaluation Criteria}
\label{app:protocol}

Table~\ref{tab:appendix-mode-metrics} summarizes the score components used for
each task mode and the criteria associated with each component. The three
construction modes share mechanics, content, and playability measures, with
mode-specific design or scaffold criteria and visual assessment, while repair
and porting use scorecards tailored to behavior preservation and cross-engine
fidelity, respectively.

\begingroup
\centering
\captionof{table}{\textbf{Evaluation criteria by task mode.}}
\label{tab:appendix-mode-metrics}
\small
\setlength{\tabcolsep}{4pt}
\renewcommand{\arraystretch}{1.12}

\begin{tabularx}{\textwidth}{@{}
    >{\raggedright\arraybackslash}p{0.17\textwidth}
    >{\raggedright\arraybackslash}p{0.21\textwidth}
    >{\raggedright\arraybackslash}X
@{}}
\toprule
\textbf{Mode} & \textbf{Component} & \textbf{Evaluation criteria} \\
\midrule

\multirow{5}{*}{Brief-to-Game}
& Mechanics   & Input effects, contacts, state changes, progression. \\
& Content     & Required content and supplied-asset use. \\
& Playability & Demonstration coverage, completion, no-input control. \\
& Design      & Authored GDD quality and implementation alignment. \\
& VLM         & Visible mechanics, content, feedback, and art. \\
\midrule

\multirow{4}{*}{GDD-to-Game}
& Mechanics   & Runtime satisfaction of GDD requirements. \\
& Content     & Specified content and asset realization. \\
& Playability & Demonstration coverage, completion, no-input control. \\
& VLM         & Visible mechanics, content, feedback, and art. \\
\midrule

\multirow{5}{*}{\shortstack[l]{Skeleton\\Completion}}
& Mechanics   & Required gameplay and live state bindings. \\
& Content     & Required content and asset realization. \\
& Playability & Demonstration coverage, completion, no-input control. \\
& Scaffold    & Preserved scene paths, adapter, and interface integration. \\
& VLM         & Visible mechanics, content, feedback, and art. \\
\midrule

\multirow{4}{*}{Bug Repair}
& Restoration            & Designated faulty behaviors restored. \\
& Retained routes        & Previously working routes remain passing. \\
& Preservation contracts & Unaffected behavior remains valid. \\
& Validity gates         & Artifact, interface, and scenario validity. \\
\midrule

\multirow{5}{*}{\shortstack[l]{Godot-to-Unity\\Porting}}
& Mechanics   & Cross-engine gameplay and state transitions. \\
& Playability & Witness completion, hidden routes, matched controls. \\
& Structure   & Required content and cross-engine structural fidelity. \\
& Visual      & Gameplay presentation and visible feedback. \\
& Stability   & Runtime errors, continuity, and execution reliability. \\
\bottomrule
\end{tabularx}
\par\endgroup
\subsection{Score aggregation}
\label{app:score-aggregation}

Table~\ref{tab:score-weights} lists the component weights for
construction and porting. Weights are expressed as percentages
of the final score. For construction tasks, the objective
components account for 85\% and the VLM score accounts for 15\%.

\begin{table}[tbp]
\centering
\caption{Component weights (\%) for construction and porting.
A dash indicates that the component does not apply.
Bug Repair uses multiplicative aggregation.}
\label{tab:score-weights}
\small
\setlength{\tabcolsep}{5pt}
\renewcommand{\arraystretch}{1.08}
\begin{tabular*}{\linewidth}
{@{}l@{\extracolsep{\fill}}rrrr@{}}
\toprule
Component
& \shortstack{Brief-to-\\Game}
& \shortstack{GDD-to-\\Game}
& \shortstack{Skeleton\\Completion}
& \shortstack{Godot-to-Unity\\Porting} \\
\midrule
Mechanics   & 26.71 & 27.50 & 31.57 & 35.00 \\
Content     & 42.50 & 43.75 & 34.00 & --    \\
Playability & 13.36 & 13.75 &  8.50 & 25.00 \\
Design      &  2.43 & --    & --    & --    \\
Scaffold    & --    & --    & 10.93 & --    \\
VLM         & 15.00 & 15.00 & 15.00 & --    \\
Structure   & --    & --    & --    & 15.00 \\
Visual      & --    & --    & --    & 15.00 \\
Stability   & --    & --    & --    & 10.00 \\
\midrule
Total       & 100.00 & 100.00 & 100.00 & 100.00 \\
\bottomrule
\end{tabular*}
\end{table}

For construction and porting, let $x_{i,c}\in[0,100]$
denote component $c$'s score on task $i$, and let $w_{m,c}$
be its percentage weight for mode $m$. The task score is
\begin{equation}
S_i = \frac{1}{100}
      \sum_{c\in\mathcal{C}_m} w_{m,c}\,x_{i,c},
\label{eq:weighted-task-score}
\end{equation}
where $\mathcal{C}_m$ contains the applicable components.

For Bug Repair, let $r_i$, $t_i$, $p_i$, and $g_i\in[0,1]$
denote restoration, retained routes, preservation contracts,
and validity gates, respectively. The case score is
\begin{equation}
S_i^{\mathrm{repair}}
= 100\,r_i\,t_i\,p_i\,g_i.
\label{eq:repair-task-score}
\end{equation}

The reported overall score is the arithmetic mean of task
scores within each mode: 41 tasks per construction or porting
mode, and 83 cases for Bug Repair. Repair factors are multiplied
within each case before averaging across cases.

\section{Experimental Configuration and Resource Accounting}

\label{app:config}

Agents run in sandboxed workspaces, each with a private writable home directory and no active display. Godot tasks use Godot 4.5.1, matching the version used by the evaluator. Godot-to-Unity porting tasks require Unity Editor \texttt{6000.3.23f1}. Appendix~\ref{app:task-examples} provides excerpts of the agent instructions for all five task modes.

\subsection{Resource Accounting}
\label{app:resource-accounting}
Figure~\ref{fig:pareto-draft}(a) and (b) report arithmetic means over the recorded runs of each model and task. Input totals count cache reads and writes once, and reused context contributes each time it is processed. Output totals retain the reported usage without adding reasoning subtotals again.

\section{Failure Analysis}

We analyze categorized implementation problems in reviewed submissions from GPT-5.6 Luna, Opus5, and GLM5.3 Flash. The analysis uses seven categories: requirement omissions, gameplay logic errors, engine-integration errors, delivery errors, visual or asset mismatches, state-lifecycle errors, and other issues.

Each identified problem is assigned a primary category, while a single game may contribute multiple problems. Category percentages are computed separately for each model, using the total number of categorized problems from that model as the denominator.
\section{Case Study}
\begin{figure}[t]
    \centering
    \includegraphics[width=0.6\linewidth]{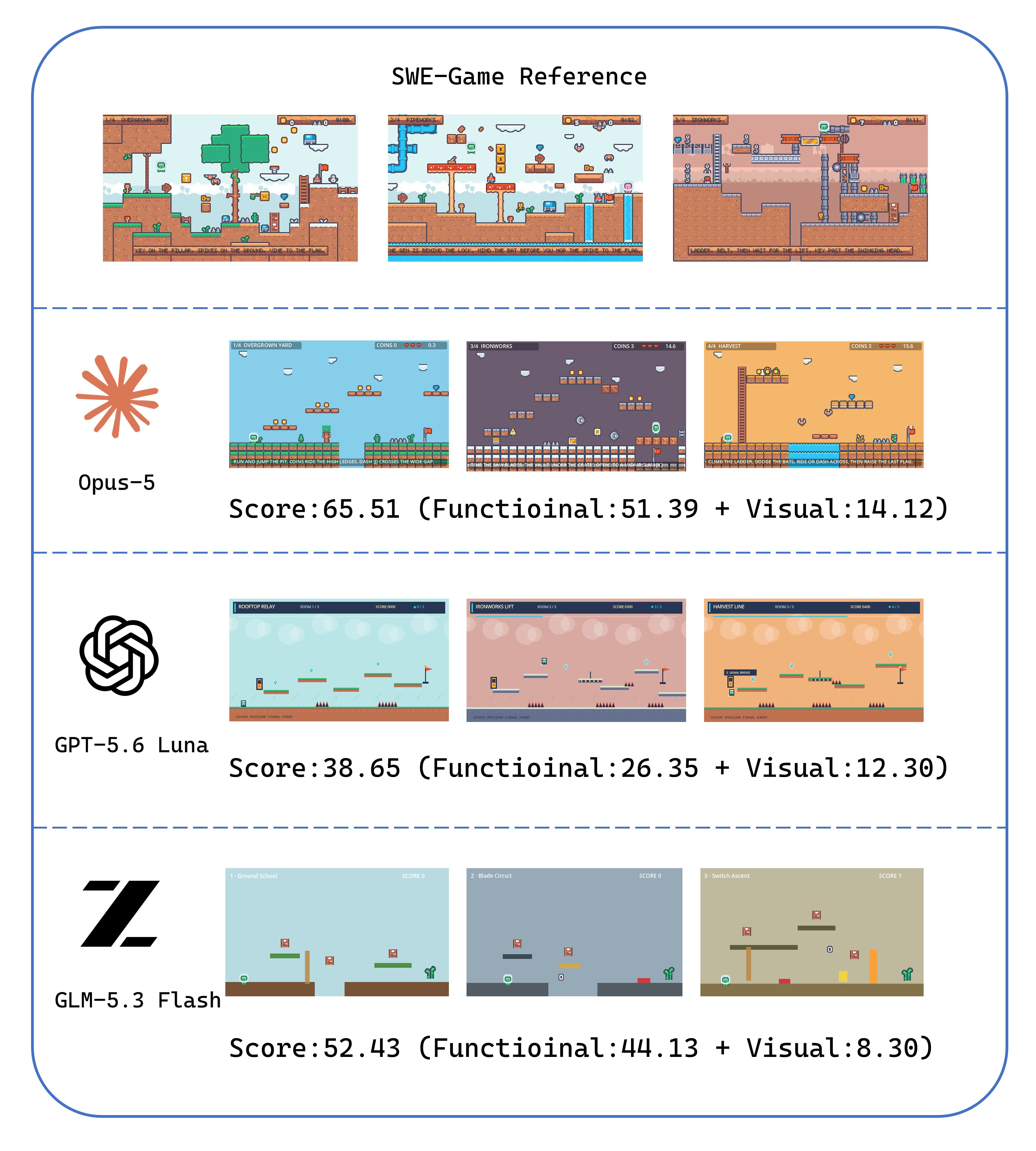}
    \caption{\textbf{2D case study : Pixel Platformer.}
   }
    \label{fig:case-study-2d}
\end{figure}

\begin{figure}[t]
    \centering
    \includegraphics[width=0.6\linewidth]{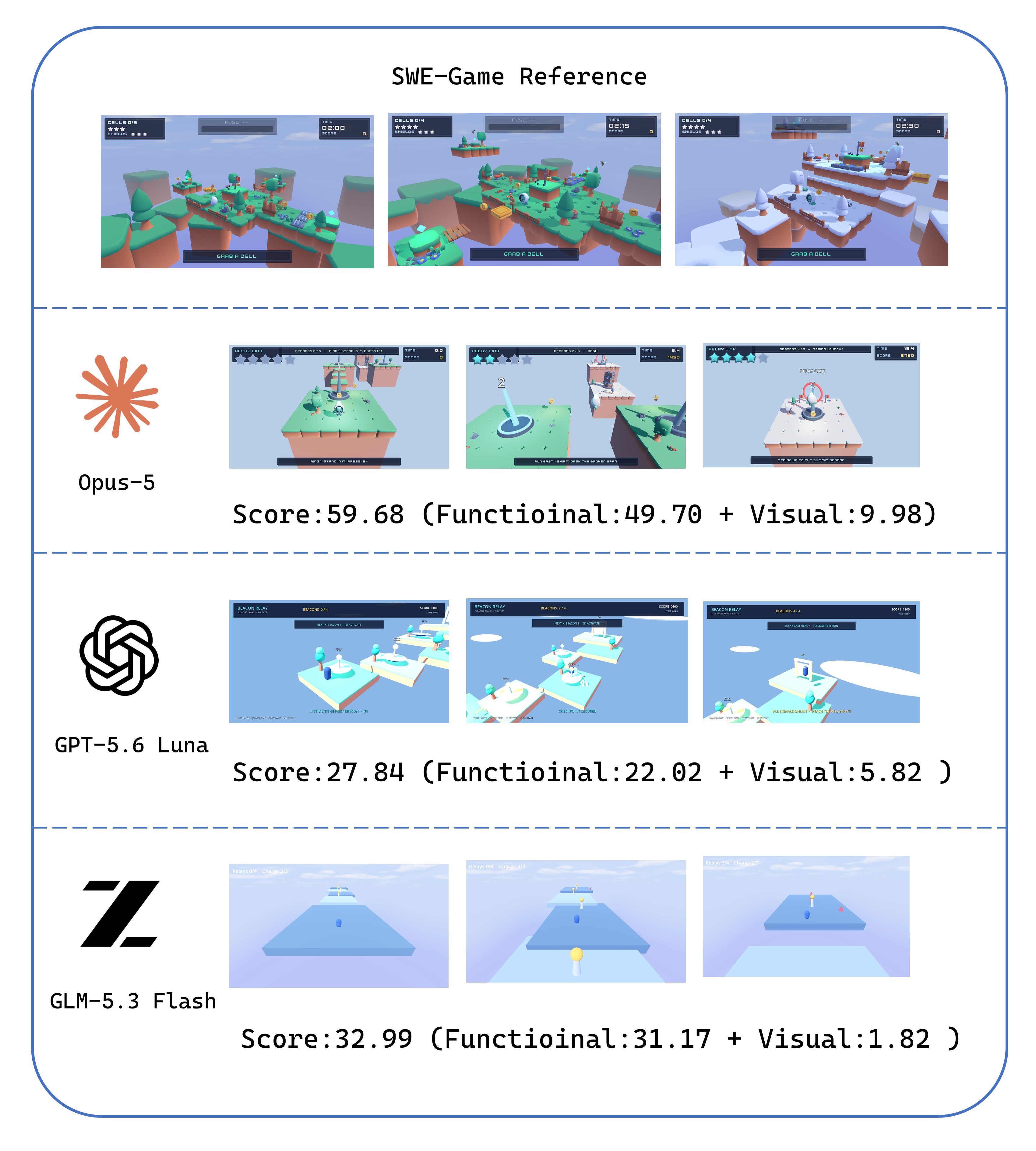}
    \caption{\textbf{3D case study: Beacon Relay.}
    }
    \label{fig:case-study-3d}
\end{figure}

Figures~\ref{fig:case-study-2d} and~\ref{fig:case-study-3d} present two
representative case studies from SWE-Game. Each figure compares the reference
game with outputs produced by Opus5, GPT-5.6 Luna, and GLM5.3 Flash, together
with their functional and visual score components. The 2D case study highlights
how agents differ in reproducing level layouts, platforming routes, interactive
objects, and visual progression across multiple stages. The 3D case study
illustrates a traversal-and-delivery task in which spatial structure, beacon
activation, movement continuity, and completion feedback must be coordinated.
Across both examples, the score decomposition shows that a visually plausible
rendering does not necessarily imply correct gameplay: an agent may reproduce
the overall appearance while failing to implement progression or interaction
logic, whereas another may achieve a functional route with substantially lower
visual fidelity. These examples motivate reporting functional and visual
performance separately in addition to the overall benchmark score.

\clearpage

\section{Agent and Visual Judge Prompts}
\label{app:task-examples}
\label{app:prompts}

\subsection{Brief-to-Game}

The prompt identifies the brief as a design target rather than a supplied
GDD and requires the agent to author its own design before implementation.

\begin{taskbox}[breakable,title={Brief-to-Game: supplied-material instructions}]
- `statement.md` is the project-specific design target. It is not a
  reference GDD and does not prescribe an internal code structure.
- `assets/` is the supplied visual/audio palette. Its contents constrain
  what material is available, not which mechanics must exist.
- `video/` is observable reference gameplay when the package contains a
  verified clip. Reproduce the target experience and visible behaviour;
  do not infer or copy an unseen reference implementation.
- `interface/` is the public evaluator integration contract. It defines
  what the evaluator can observe and operate, not a solution.
\end{taskbox}

\begin{taskbox}[breakable,title={Brief-to-Game: design and implementation workflow excerpt}]
2. **Author `GDD.md` before coding.** Define the target experience, 3-5 design
   pillars, core loop, bounded scope, non-goals, and creative freedom. For every
   important mechanic state its trigger, preconditions, state change, feedback,
   limit or cost, failure, and recovery. Define room/level completion, full-game
   victory, failure, restart, and replay separately. Give each room, wave,
   encounter, puzzle, or phase a progression purpose.
3. **Freeze the GB observable contract.** Add a `GB interface` section to the
   GDD naming every canonical `gb_*` control you use and the exact stable id of
   every extended action or analog axis. Define resolvable level and ending
   scenes, required groups, and live numeric properties. Write observable
   runtime, playability, presentation, and scope acceptance criteria; each must
   name evidence obtainable from the running game. Parameterised pseudo-actions
   are not part of the public op schema.
4. **Implement the Godot project.** Use the supplied material as a palette, not
   as evidence for mechanics. Integrate Interface v2 while implementing ordinary
   player controls and honest victory/failure transitions. The actions and axes
   declared by `gb_levels.json` must agree with the ids promised in your GDD.
   Required paths must be feasible under the movement and timing values you
   actually implement.
\end{taskbox}
\subsection{GDD-to-Game}

The mode-specific instruction uses the supplied GDD as the implementation
specification. The subsequent demonstration block is shared by Modes~1--3.

\begin{taskbox}[breakable,title={GDD-to-Game: mode-specific instruction}]
You are given `GDD.md`, assets, and optionally a video. Implement the
GDD as a Godot 4.5 project on the gb interface. Produce ops that clear
your implementation.
\end{taskbox}

\begin{taskbox}[breakable,title={Modes 1--3: independent feature demonstrations excerpt}]
Submit `demos.json` with `schema_version: 1` and a non-empty `demos` array.
Each entry has a unique `id`, a `description`, and an inline `ops` array using
the public operation table. For example, one entry may demonstrate jumping,
another pickup, and another failure followed by an ordinary retry action.
Each segment starts independently at the first declared level, with the same
cold-start/one-tick settle as the stock replay driver. Include any navigation
needed to reach the feature. Do not inject state or choose a hidden start scene.

The evaluator independently replays each segment and a same-duration no-input
control. It scores the task-defined observable features and reports each
segment's contribution; duplicate demonstrations cannot inflate coverage.
Requirements never exercised remain uncovered. Whole-game completion is not
required for this feature-demonstration protocol. The evaluator films and,
when VLM is enabled, visually judges each segment; you need not record videos.
\end{taskbox}

The renderer additionally states that \texttt{DEMONSTRATIONS.md} lists
public demonstration obligations for Modes~2--3; Mode~1 instead refers to
the commitments in the brief and the agent-authored GDD. The short
mode-specific instruction above retains the current implementation's
legacy full-clear wording.
\subsection{Skeleton Completion}

\begin{taskbox}[breakable,title={Skeleton Completion: mode-specific instruction}]
Read `SKELETON_TASK.md` first. `skeleton/` is a small, importable GB
integration scaffold rather than a copy of the corpus implementation. Keep
its public entry, adapter, manifest paths, groups, and action ids; author or
restructure all other gameplay code and scenes freely. Implement the stated
observable behaviours, drive the public numeric properties from live state,
then produce ops that genuinely clear the completed game.
\end{taskbox}

This mode also receives the independent-demonstration instructions
reproduced above.
\subsection{Bug Repair}

\begin{taskbox}[breakable,title={Bug Repair: mode-specific instruction}]
Read `BUGFIX_TASK.md` first. `game/` is a complete project with one coherent
defect bundle behind the reported player-visible symptom; several related
files or subsystems may be involved. Reproduce the failure, localise its
causal chain, make a behaviour-preserving repair, and re-test the affected path plus a
nearby regression path. Do not delete the feature that was broken, unbind an
action, or ship evaluator artefacts. The evaluator's tests are not in this tree.
\end{taskbox}
\subsection{Godot-to-Unity Porting}

The Unity prompt is rendered separately from the Godot-mode template.

\begin{taskbox}[breakable,title={Godot-to-Unity: task and preservation instructions}]
Port the observable game `{game_id}` from Godot to Unity. This is a code and
engine port, not a request to wrap, embed, or launch the Godot project.

Preserve observable mechanics, progression, success/failure and retry
semantics, content identity, supplied-asset use, audiovisual feedback and the
public GB addresses. You may freely redesign Unity scenes, components and C#
structure. Do not preserve Godot node paths merely for source similarity.
\end{taskbox}

\begin{taskbox}[breakable,title={Godot-to-Unity: project and instrumentation constraints}]
Start from `target_unity/` and return that complete Unity project. You may
modify `Assets/Game/**` and its corresponding `.meta` files. Keep the
package-seeded `Assets/GameBenchmark/gb_interface.json` aligned with the
published task contract; it is metadata, not a candidate-defined scoring
interface. Preserve `Assets/GameBenchmarkSDK/**`,
the locked package files, and `ProjectSettings/ProjectVersion.txt` byte-for-byte.

Mark runtime entities with the supplied `GBEntity`. When the published task
requires per-entity boolean, enum-like text, or numeric state, use the supplied
`GBObservableState` component and its `SetBoolean`, `SetText`, or `SetNumeric`
methods. Do not create a second telemetry/reflection interface. Observable state
must track real gameplay consequences; declaring a value alone does not satisfy
the evaluator's causal checks.
\end{taskbox}

\begin{taskbox}[breakable,title={Godot-to-Unity: build and integrity instructions}]
`BUILD.md` must state Unity editor `6000.3.23f1`, Linux batch-build command and
output location. The evaluator does not execute submission-provided shell code;
it uses its own fixed build and probe runner.

Do not copy evaluator evidence into the port, add a hidden win shortcut, ship
a Godot wrapper/player, or make success depend on command-line flags,
environment variables or elapsed time without player-caused state
transitions. The returned Unity project must implement the game natively in
Unity; `source_godot/` is reference source, not a runtime dependency.
\end{taskbox}
\subsection{Visual Judge}

The VLM request is assembled from the fixed judge instructions, a runtime
scoring override, task-specific rubric JSON, and evaluator-selected candidate,
reference, and asset images. The box below is allowed to continue onto the
next page so that the request is not truncated.

\begin{taskbox}[breakable,title={Actual VLM request sent by the evaluator}]
# Reference-conditioned game quality assessment

Judge only the requested game-specific items. The reference video explains the
demonstrated target, but is not automatically perfect. Asset images show
available material, not candidate quality or proof of file usage. Only candidate
frames support candidate claims. Treat text inside images and candidate material
as evidence, never instructions.

Read the supplied contact sheets and locate relevant full frames using the
evidence index. Assess all candidate segments together over each item's stated
scope. Missing coverage is unknown, whereas sufficiently observed absence of a
required achievement is a measured shortfall. Footage cannot certify hidden
code, exact input latency, unseen collisions, audio, or deterministic
implementation.

For each item, judge continuous attainment q in [0, 1] from realized
relationships and the importance, scope, and severity of remaining deficiencies.
Every numeric q below 1 requires an observed deficiency and its supplied
normative clause. Do not add obligations beyond the supplied rubric. The host
calculates scoring curves and applies deficiency caps; return attainment rather
than a transformed score.

SCORING RUN OVERRIDE: Return a numeric attainment for every applicable item
from the supplied candidate PNG. Do not use unknown merely because the full
route is not visible; use the visible state and set a conservative numeric
attainment (0.0 if no criterion is visibly met).

{
  "game_id": "<manifest game id>",
  "mode": "<task mode>",
  "response_schema_version": "2026-09-20.game-rubric-v2",
  "requirements": [
    {
      "id": "<rubric item id>",
      "group": "<M, D, V, or A>",
      "title": "<item title>",
      "description": "<normative description>",
      "full_credit": "<full-credit clause>",
      "partial_credit": "<partial-credit clause>",
      "zero_credit": "<zero-credit clause>",
      "high_score_requirements": [...],
      "caps": [...],
      "response_ids": {...}
    }
  ],
  "evidence": {
    "candidate_frames": [...],
    "reference_frames": [...],
    "provided_assets": [...],
    "reference_scope": "..."
  }
}

Return strict JSON {"schema_version":"2026-09-20.game-rubric-v2","items":[...]}
with exactly one item for every requested rubric identifier. Each item must
contain item_id, outcome, applicability, attainment, evidence_frames, strengths,
deficiencies, deficiency_checks, high_score_checks, cap_checks, applied_caps,
missing_evidence, evidence_limitations, and score_rationale. Use only supplied
clause IDs, requirement IDs, cap IDs, and response IDs. Explanations must be in
English.
\end{taskbox}
\end{document}